\documentclass{sig-alternate-05-2015}
\usepackage{enumitem}
\usepackage{multirow} 
\usepackage{hhline}
\usepackage{gensymb}
\usepackage{algorithm2e}
\usepackage{amssymb}
\usepackage{subcaption}

\begin{document}

\setcopyright{acmcopyright}

\title{Unsupervised Method to Localize Masses in Mammograms}
%
%
%
%
%

\numberofauthors{1} 
%
\author{
%
%
\alignauthor
Bilal Ahmed Lodhi\titlenote{PhD candidate in Korea University}\\
       \affaddr{Korea University}\\
       \affaddr{Seoul}\\
       \affaddr{Republic of Korea}\\
       \email{balodhi@gmail.com}
}



\maketitle
\begin{abstract}
Breast cancer is one of the most common and prevalent type of cancer that mainly affects the women population. chances of effective treatment increases with early diagnosis. Mammography is considered one of the effective and proven techniques for early diagnosis of breast cancer. Tissues around masses look identical in mammogram, which makes automatic detection process a very challenging task. They are indistinguishable from the surrounding parenchyma. In this paper, we present an efficient and automated approach to segment masses in mammograms. The proposed method uses hierarchical clustering to isolate the salient area, and then features are extracted to reject false detection. We applied our method on two popular publicly available datasets (mini-MIAS and DDSM). A total of 56 images from mini-mias database, and 76 images from DDSM were randomly selected . Results are explained in-terms of ROC (Receiver Operating Characteristics) curves and compared with the other techniques. Experimental results demonstrate the efficiency and advantages of the proposed system in automatic mass identification in mammograms.

\end{abstract}

%
%


%
%

%
%



\section{Introduction}
\label{Introduction}
Breast cancer is the most common cause of cancer-related deaths among women worldwide. With more than 450, 000 deaths each year, breast cancer accounts for about 14\% of all female cancer deaths (\cite{law2014automated}). Recent statistics says that 1 out of 10 women is affected by breast cancer in their lifetime. According to GLOBOCAN 2012, 1.7 million Women were diagnosed with breast cancer and there were 6.3 million women alive who had been diagnosed with breast cancer in the previous five years (\cite{brigham}).  Although the breast cancer rate is increasing in many  parts of the world, however the mortality rate is much higher in less developed countries, because of insufficient facilities available for diagnosis and treatment. Therefore, there is an urgent need of reliable and affordable approches for early diagnosis and treatment of breast cancer in less developed countries.It can have significant impact on cancer treatment, faster recovery and reducing mortality. 
\par Mammography is considered most effective technique as it can detect 85$\sim$90\% percent of all breast cancers (\cite{brigham}).  A mass is an uncontrolled grown tumor and we classify them into malignant and benign by their size, shape and other features. As described earlier that early diagnosis is a key for effective treatment. Therefore the job radiologist becomes very important, who can interpret mammograms for early diagnosis. Mammogram does not have so much information imprinted on the film. Cancer diagnosis in this scenario becomes a subjective criteria. Radiologist opinion depends on their experience. \cite{surendiran2008soft} states that radiologist's diagnosis inter-observer variation rate is 65 $\sim$ 75\% ). He can miss a significant proportion of abnormalities and in addition a large number of mass come out to be benign after biopsy (\cite{surendiran2008soft}).  \cite{lesniak2011computer} states that Computer aided diagnosis (CAD) systems is helpful for the radiologists in diagnosis. (\cite{wei2005study}) claims that detection accuracy improved by combining the expert knowledge with CAD scheme.We proposed an algorithm to address the previously described problem for breast cancer diagnosis. Proposed scheme is novel in the following ways:
\begin{itemize}
\item Scope of the detection algorithm is wide. It can detect different type of cancers in malignant and benign categories. Proposed algorithm was tested on many ill-defined masses also.
\item A method is proposed to identify masses, irrespective of their size and shape.
\item We proposed an efficient and unsupervised approach to detect masses in mammogram images. It segments the breast region and finds the candidate regions of interests  (ROIs).
\item Generalization of algorithm is tested by experimenting cross validation across two different datasets. 
\end{itemize}

\par The organization of paper is as follows.  Section I presents introduction and significance of the work. Section II discusses previous and related work. Section III briefly describes the proposed method for preprocessing. Section IV analyses the results and finally, Section V concludes the article.

\section{Related Work}
\label{Related Work}
In order to develop computer aided breast cancer detection tools, researchers have used several approaches. \cite{kuo2014mass} proposes a Particle Swarm Optimized Wavelet Neural Network (PSOWNN) based classification approach for detection of masses in digital mammograms. Their method is based on extracting Laws Texture Energy Measures from the mammograms and classifies the suspicious regions by PSOWNN. Their method does not have any noise removal algorithm and also they do not propose any intelligent method ROI detection. In (\cite{wang2014latent}, \cite{vallez2014breast},  authors used Latent Dirichlet Allocation (LDA) to mine the feature set of mammogram images. They presented the modified Morphological Component Analysis method to identify the mass region and then extracted morphological features. Finally, LDA is used to classify the masses. Simple Morphological approaches are sensitive to noise. They also did not presented any preprocessing for collection of ROIs.
\par In \cite{sun2014prediction}, authors proposed the modified Fuzzy c-means clustering to cluster the masses, extracted morphological, textual and spatial features and classified the features using SVM (Support Vector Machine). Their method lacks the noise removal and intelligent ROI segmentation.  \cite{pereira2014segmentation} presented a set of tools to aid segmentation and detection of mammograms that contained mass. After the top-hat morphological operator, de-noising is applied. Image gray-level was enhanced by wavelet transform and wiener filter. Finally, segmentation method was employed using multiple thresholding, wavelet transform and genetic algorithm. They used manual process to reduce the false positives generated by genetic algorithm. Authors also did not do the automatic classification of the ROIs.  \cite{agrawal2014saliency} proposed a method for mass detection based on saliency map. After the creation of saliency map, a threshold is used to obtain the ROI. A number of features were extracted and classified by SVM.  Automated detection of malignant masses in screening mammography has been discussed in \cite{rodriguez2013detection}. It developed a technique that used presence of concentric layers which surrounds a focal area in the breast region, that has suspicious morphological characteristics and low relative incidence. Segmentation process in both of the earlier described algorithms are focused on the bright or salient parts of the image, which is always mis-leaded by the blood vessels resulting in the whole breast parenchyma as a ROI.   \cite{mavroforakis2005significance} work is based on applying one-dimensional recursive median filter to different number of angles to each pixel. It becomes difficult to detect when structure of the mass and a normal glandular looks similar. It can only be detected if there were asymmetry between the left and right breasts.
\par \cite{mudigonda2001detection} proposed method is based on the analysis of ISO-intensity contour groups to segment skeptical masses.  False positives are then removed using features based on flow orientation in adaptive ribbons of pixels across the margins of masses. The procedure is tested on 56 images from the Mini-MIAS database and got a sensitivity at the rate of 81\% at 2.2 false positives per image. Furthermore, based on gray-level co-occurrence matrices (GCM) and using features on a logistic regression method, the classification of masses were performed as benign or malignant using five texture features. An accuracy of 0.79 is achieved as a result of this classification, with 19 benign and 13 malignant lesions. Authors used the  hard thresholds to get the contours of objects in the image. Contour are very sensitive to noise resulting in increase of false positives and bad segmentation. Algorithm will fail to detect the mass if the boundary is ill-defined or even the mammogram is much denser.
\par \cite{eltoukhy2014optimized} proposed a method for diagnosis of breast lesions (diagnosis). Masses using the wavelet transform to obtain a multi-resolution representation of the original image at each resolution, a set of features is extracted which serves as input to a binary tree classifier. Algorithm achieved 91.9\% true positive detection accuracy.  ROIs were manually cropped in the proposed system. Their proposed system is based on wavelet and curvelet coefficients, which is very high in numbers. Selecting best coefficients is an optimization problem and also it is very sensitive to noise. \cite{zheng1999detection} proposed a method combines several artificial intelligence techniques with the discrete wavelet transform (DWT). ROI's are determined through dimensional analysis using a multi-resolution Markov random field algorithm, the segmentation is performed that leads to the application of tree type classification strategy. The algorithm was tested in the Mini-MIAS database and has a sensitivity of 97.3\% with 3.9 false positives per image. Their proposed method works well with well-defined masses, but ill-defined masses are difficult to be classified by this method.

\section{Methodology}
\label{Methodology}
\begin{figure}
\centering
\includegraphics[ scale=0.45]{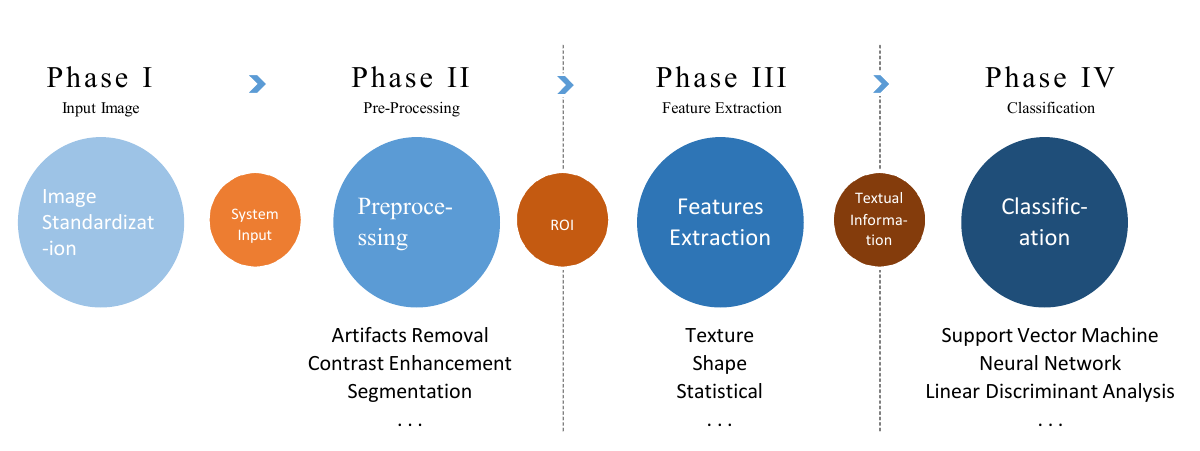}
\caption{General phases of Detection Algorithm \label{fig:gphases}}
\end{figure}

Female breast parenchyma  is a multiplex biological structure and is composed of glandular, fatty, and lymphatic tissues (lymphovascular structures). Mammography imprints the texture information of breast tissue in image. Though the composing components may be complicated, the mass regions are characterized of high intensity and high texture. Figure \ref{fig:gphases} shows the process of a typical analysis system.
\par
\begin{figure}
\centering
\includegraphics[ scale=0.45]{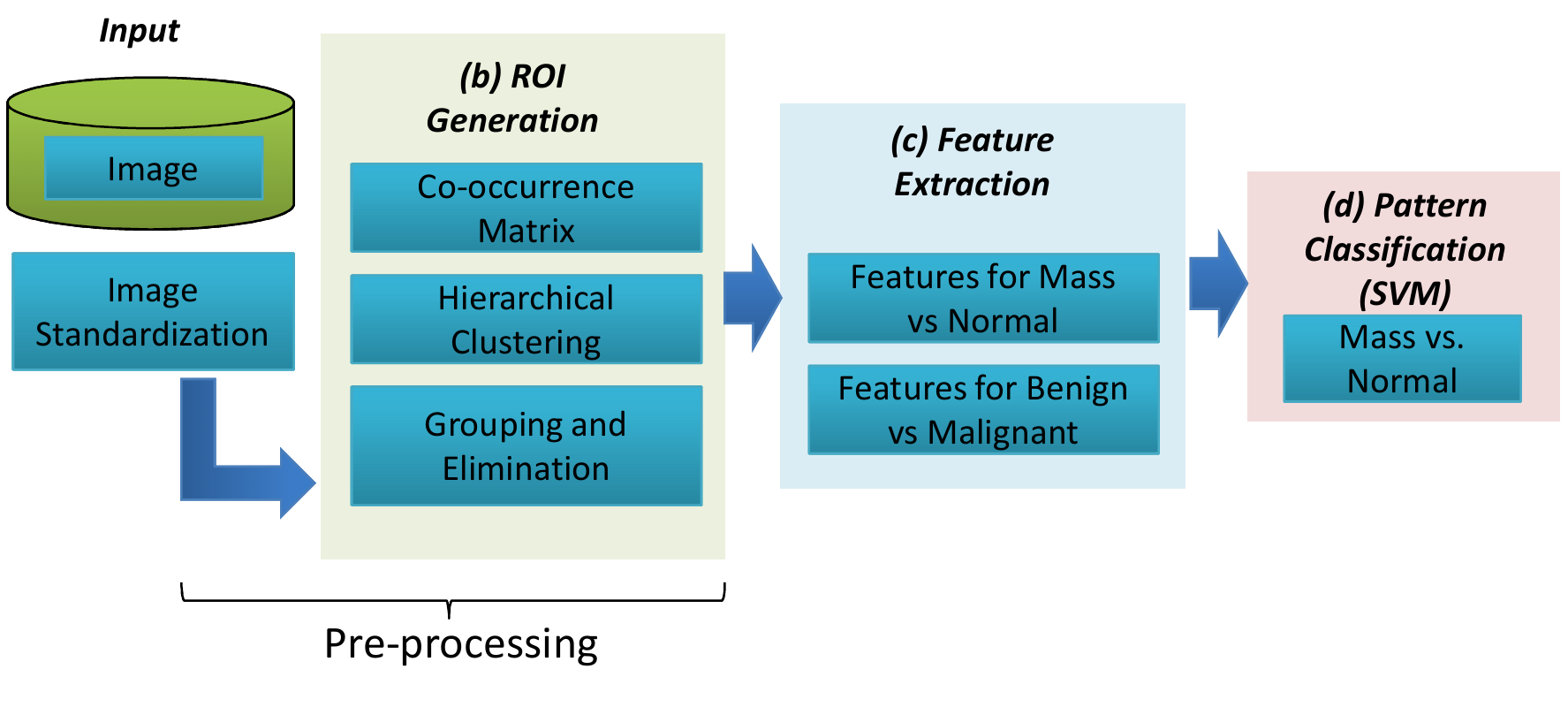}
\caption{Overview of the Proposed Algorithm \label{fig:abstractView}}
\end{figure}

 We propose an efficient and unsupervised approach to identify the suspicious regions in mammogram images. Proposed algorithm isolates the spatially interconnected structures in the image, which are concentrated around salient intensities. As a result, it is possible to extract high-level information to analyze further, to characterize the physical properties of mass regions and to prepare a short-list of skeptical ROIs. Figure \ref{fig:abstractView} shows our proposed algorithm. Further explanation of the algorithm is explained in the following subsections.
\subsection{Image Standardization}
Data from different sources should be converted to one format. Proposed algorithm was tested on two datasets: Digital Database for Screening Mammography (DDSM) (\cite{heath1998current}:\cite{heath2000digital}) and Mammographic Image Analysis Society Database (mini-MIAS) (\cite{suckling1994mammographic}). MIAS dataset is in Portable Gray Map (PGM) format while DDSM dataset contains images in LJPEG format. We converted the DDSM dataset TO 16-bit Portable Network Graphics (PNG) format by a wrapper program developed by us \footnote{Utilities at http://microserf.org.uk/academic/Software.html were used to write a wrapper program.}.

\subsection{ROI Detection Phase}
\label{ROI Detection}
One of the main tasks is to get mass-candidate regions. Following subsections describe the way to get those regions.
\subsubsection{Smoothing}   It is assumed that malignant masses typically cause distortion to the surrounding tissues. So, segmentation process can over-segment the image and it can't get those masses in a single entity. To overcome this problem, prior smoothing of the image is necessary. In the present work, Gaussian pyramid is used to uniformly highlight the salient regions. Subsampling to many levels results in over smoothing the image which converts the image regions as blobs. However, some researchers (\cite{petrick1996adaptive}) have performed mass detection on reduced resolutions of 800m.
\begin{figure}
\centering
\includegraphics[scale=0.2]{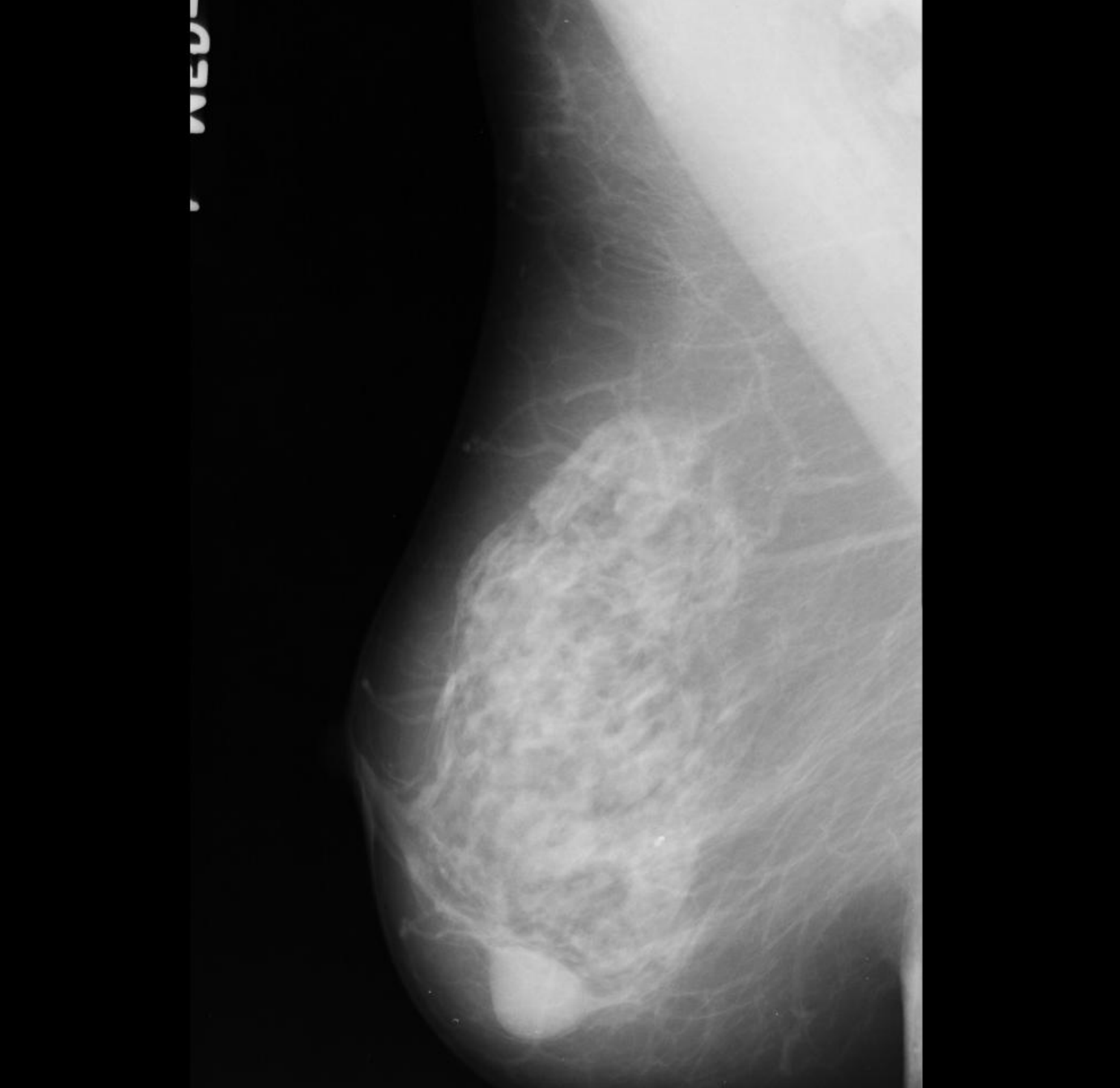}
\caption{Original Image from MIAS dataset \label{fig:orignalCase}}

\end{figure}
Regions of mass are hyper-densed. We need to get the full mass area to extract meaningful features from the ROI.  Abrupt changes in the intensity of the objects present in the image effect the segmentation process. Peaks in the image objects are smoothed by the above described preprocessing.  

\subsubsection{Hierarchical Clustering with GLCM (Gray level Co-occurrence Matrix) data}
\label{Hierarchical Clustering}
\begin{figure*}
\centering
\includegraphics[scale=0.4]{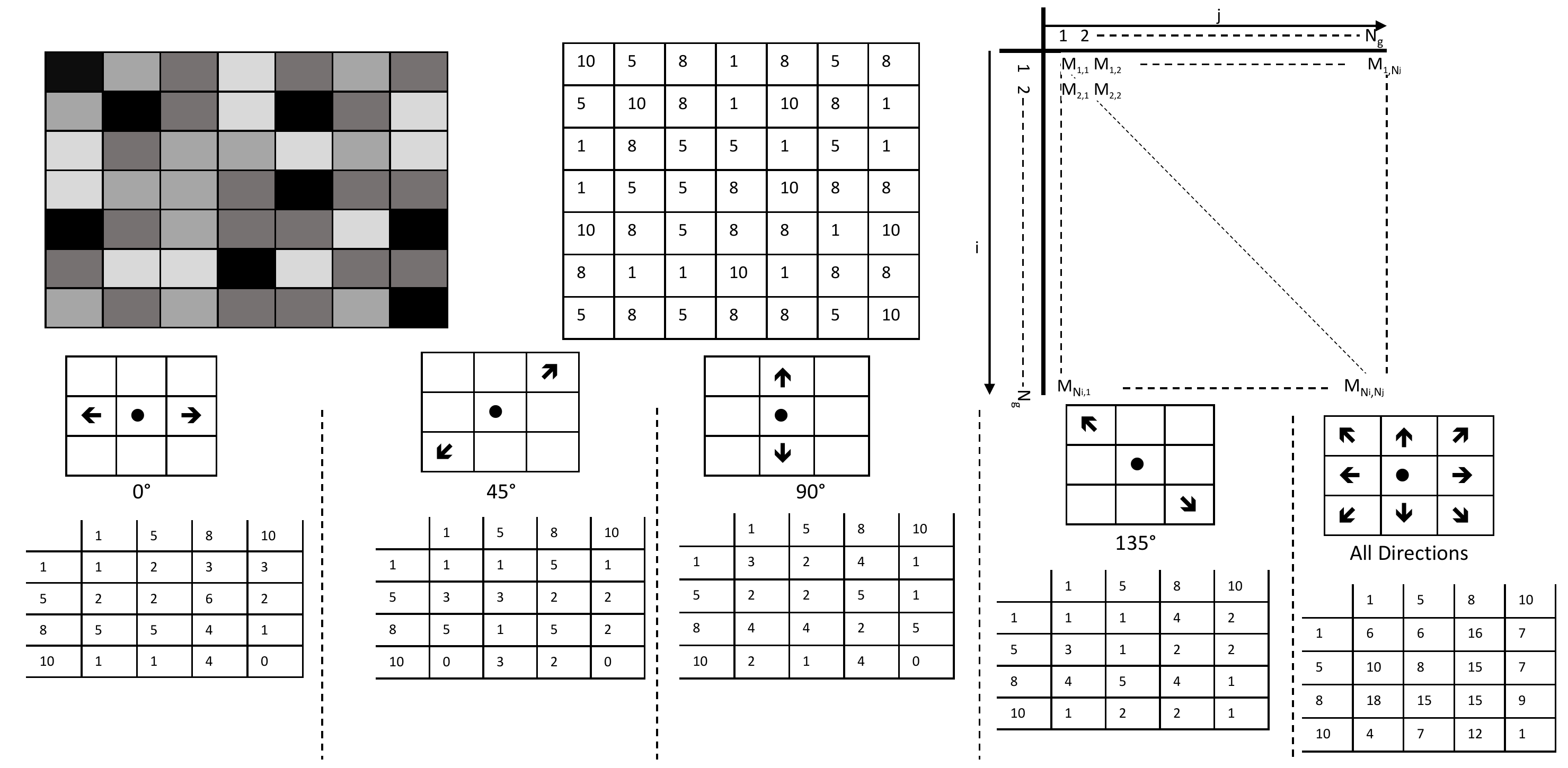}
\caption{Process of co-occurrence matrix \label{fig:comat}}
\end{figure*}

We applied hierarchical clustering with GLCM data to segment the salient regions of image.  Before segmentation of the image, its contrast was enhanced by CLAHE (Contrast Limited Adaptive Histogram Equalization). Further, we calculate the gray-level co-occurrence matrix from image. GLCM is created with distance one and 4 directions [0 1; -1 1; -1 0; -1 -1] (0$\degree$, 45$\degree$, 90$\degree$, 135$\degree$). Other angles were not computed due to redundancy of the data. GLCM data from all directions are summed up and normalized. Figure \ref{fig:comat} depicts the explanation of co-occurrence matrix.
Intensities in mass exhibit the glowing effect (intensities are propagated from the center of the masses). Hierarchical clustering can  cluster image according to propagated intensities while having a family structure of concentric objects. At each hierarchical level a measure of dissimilarity is defined to differentiate clusters and object are merged together as one, if their dissimilarity is less than or equal to the acceptable dissimilarity measure. 
\par Many researchers have proposed methods for multilevel thresholding by discriminant analysis (\cite{otsu1975threshold}, \cite{rodrigues2015two}, and \cite{arifin2004image}). They thresholded the image by the cluster analysis irrespective of the physical location of the cluster. This idea works better if the image is multi-modal and we divide it into two clusters (background and foreground). However, it does not give fine results on low-level x-rays images which are mostly unimodal. In this case, multi-thresholding does not give compact objects for ROI. We incorporated the discriminant analysis (\cite{arifin2004image}) with GLCM data to get compact objects.
The proposed method clusters the image intensities in a hierarchy, according to their co-occurrence and similarity measure. Number of thresholds are found by cutting the dendrogram at desired level.Initially, each gray-level is designated to a different cluster i.e. g gray-levels in image will generate q number of clusters and each cluster has its own threshold $T_i$ . Family hierarchy of clustering process can be viewed as a dendrogram. The estimated thresholds for the image to segment can be obtained by cutting the branch in dendrogram. Clustering algorithm is defined in algorithm \ref{alg:clustering}.

\begin{algorithm}
\SetAlgoLined
\KwResult{Return n thresholds}
Given:
A set of gray-levels \{$x_1,x_2,....,x_q$\}\;
A distance function dist(c1,c2)\;
m number of threshold levels\;
\For{i=1 to q}{
	$c_i$ = \{$x_i$\}\;
	$t_i$ = \{$x_i$\}\;
}
C = \{$c_1$,.....,$c_q$\}\;
T = \{$t_1$,.....,$t_q$\}\;
\For{k=1 to q-m}{
	- make adjacent cluster pairs\;
	- ( $c_{min1}$,$c_{min2}$)\ = minimum dist($c_i$,$c_j$) for all $c_i$,$c_j$ in C\;
	- remove $c_{min1}$ and $c_{min2}$ from C\;
	- remove $t_{min1}$ and $c_{min2}$ from T\;
	- add \{$c_{min1}$,$c_{min2}$\} to C\;
	- add \{$t_{min1}$,$c_{min2}$\} to T\;
}
\caption{Clustering Algorithm}
\label{alg:clustering}
\end{algorithm}

\paragraph{Distance Metric}
The distance measure between two clusters in the proposed algorithm is defined as ratio between the measure of observed dispersement and the expected dispersement. it is calculated as: 

\begin{equation}
\label{eq:disteq}
dist_{(q_{i},q_{j})}\ =\ {{(1-CP{_{q_{i}q_{j}}})(P_{q_{i}}-P_{q_{j}})^2{\left[{\overline{X_{q_{i}}}-\overline{X_{q_{j}}}}\right]}^{2}}\over{\sigma_{q_{i}q_{j}}^{2}}}
\end{equation}

where $q$ is the total number of clusters, $P_q$ is probability density function of image histogram and it can be calculated as equation \ref{eq:pdf}.  $CP_{i,j}$ represents the normalized co-occurrence frequency of the cluster pair being merged. It is defined in equation \ref{eq:clusterCo}. $\overline{X}$ is the mean value of the cluster and defined in equation \ref{eq:clustermean}. $\sigma^2$ is the variance of both clusters which are being merged. It is defined in equation \ref{eq:variance}.

\begin{equation}
\label{eq:pdf}
P_{q}\  =\ \sum\limits_{l=T_{q-1}+1}^{T_{q}}{h(l)}
\end{equation}
\par where $l$ represents the gray-level in image (value: [0 255])such that
$\sum\limits_{i=1}^{q}{P_{i}}=1$.    
\begin{equation}
\label{eq:clusterCo}
CP_{i,j}=\sum\limits_{t=T_{q_{j-1}+1}}^{T_{q_{j}}}{{{\sum\limits_{s=Tq_{i-1}+1}^{T_{qi}}{CM_{s,t}}}\over{T_{qi}-T_{q_{i-1}}}}}
\end{equation}
where $CM_{s,t}$ is the co-occurrence probability of gray-level s and t.

Mean is also called as the expectation of the cluster and can be represented as:
\begin{equation}
\label{eq:expectation}
\mu=E(q_{i})=\sum\limits_{i=1}^{q}{l_{i}P(l_{i})}
\end{equation}
so we calculated the mean as: 
\begin{equation}
\label{eq:clustermean}
\overline{X}_{q}\ ={{1}\over{P_{q}}}\sum\limits_{l=T_{q_{1}}+1}^{T_{q}}{lh(l)}
\end{equation}
Variance of the distribution is defined as:
\begin{equation}
\label{eq:expecvariance}
\sigma^{2}=\sum\limits_{i=1}^{q}{{\left({l_{i}-\mu}\right)}^{2}P(l_{i})}
\end{equation}
This formulates the variance into the following equation.
\begin{equation}
\label{eq:variance}
\sigma_{q_{i}q_{j}}^{2}\ =\ {\sum\limits_{l=T_{q-1}+1}^{T_{q2}}{\left[{l-\overline{CX}_{q_{i}q_{j}}}\right]}}^{2}h(l)
\end{equation}

where $\overline{CX}$ is defined as average mean of the cluster pair. It is calculated as the weighted average between the cluster means of the pair being merged:

\begin{equation}
\overline{CX}_{qiqj}\ =\ {{P_{qi}\overline{X}_{qi}\ +\ P_{qj}\overline{X}_{qj}}\over{P_{qi}+P_{qj}}}
\end{equation}

We imposed a restriction that only the adjacent clusters are allowed to merge. The similarity measurement is adapted by \cite{otsu1975threshold}. Pair having the minimum distance value is the best candidate to merge.
\par The saliency of a region is measured by the nesting depth of hierarchical clustering which identifies nested objects. One statistical parameter LevelParameter is introduced that represents the levels in hierarchical clustering. LevelParameter value of 5 is used in the study. Figure \ref{fig:foundClusters} shows the number of objects found in mammogram by segmentation process.

\subsubsection{Grouping and Elimination}
\label{Grouping and Elimination}
\begin{figure*}
\begin{subfigure}{.5\textwidth}
\centering
\includegraphics[scale=0.2]{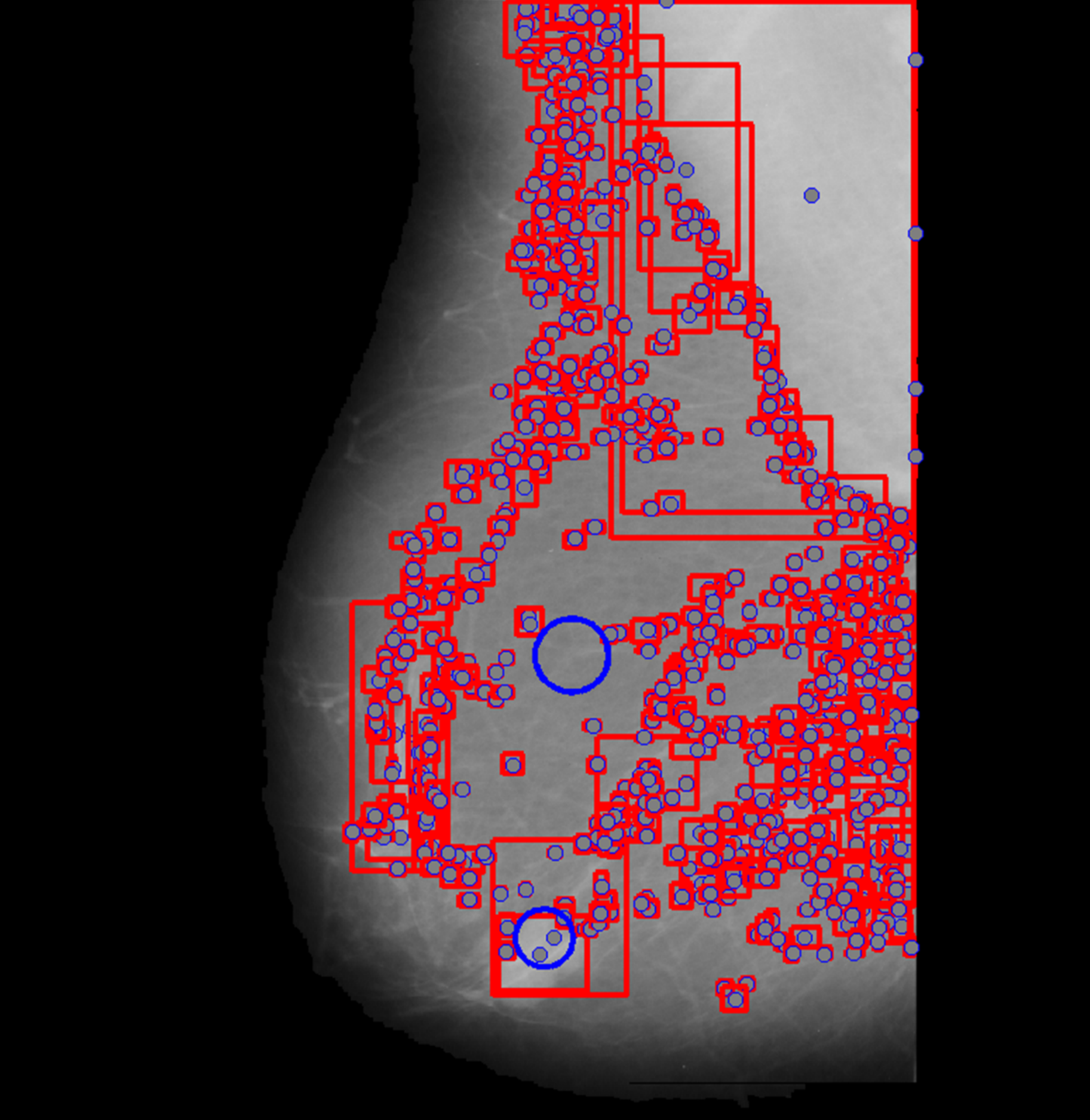}
\caption{Detected ROIs (Objects) \label{fig:foundClusters}}
\end{subfigure}
\begin{subfigure}{.5\textwidth}
\centering
\includegraphics[scale=0.2]{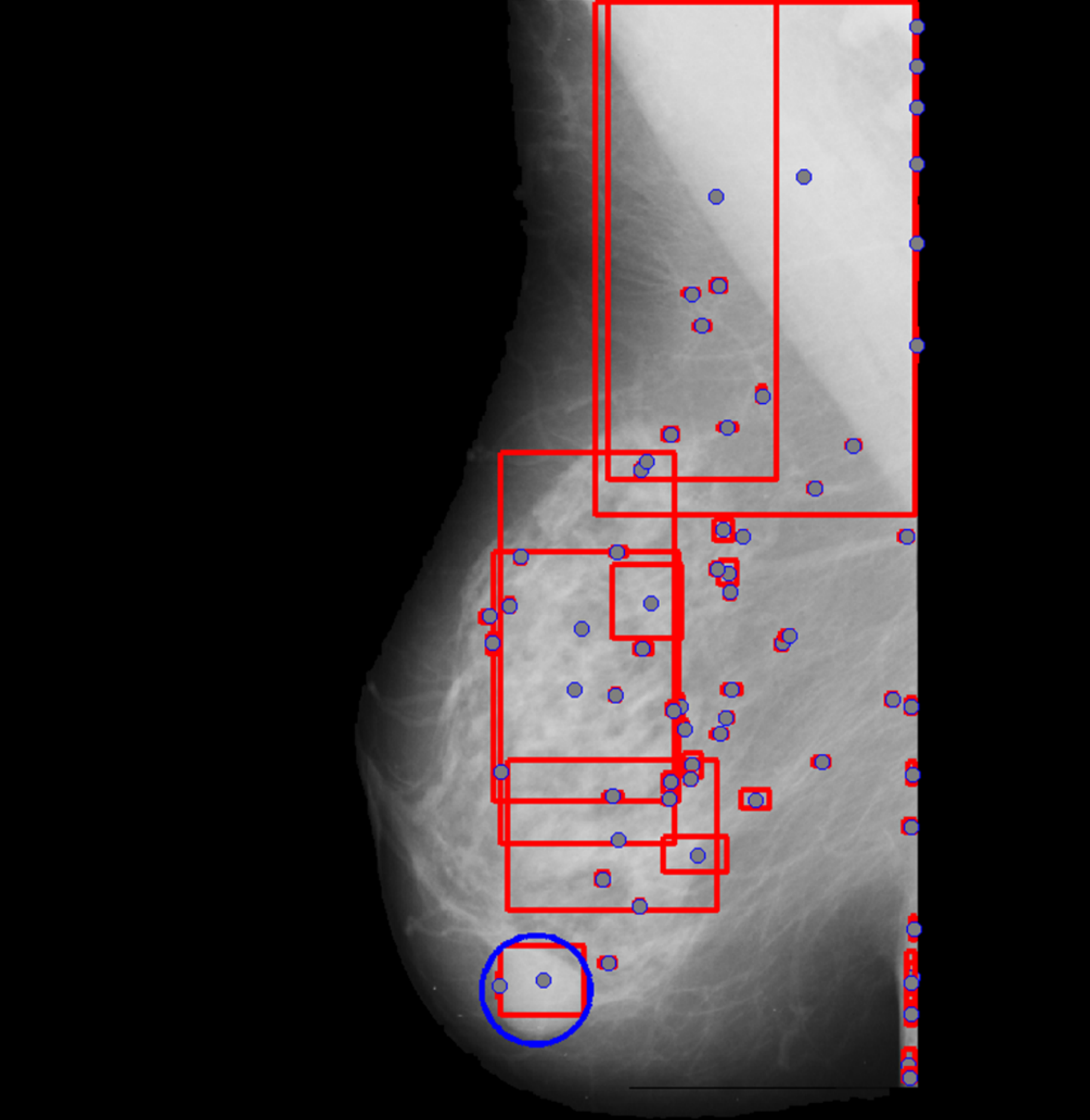}
\caption{Merged ROIs (Objects) \label{fig:mergeClusters}}
\end{subfigure}
\caption{Detected objects and their merging process}
\label{fig:detectedObjects}

\end{figure*}

Segmentation process described in previous section results in a large number segmented objects. 
We devised an algorithm to reduce the number of objects and extract only the relevant data for analysis. Our first step in this process is grouping and elimination. As previously described, masses exhibits the glowing effect, therefore, we first find the dense-core portions and then go to the next threshold level to find objects which encircles the previously detected object. The idea of prestige in link analysis is used and also the hierarchical clustering nodal relation is considered. Every possible regions are given a prestige score of 1. When they are encircled by other immediate lower density parent they forward their prestige score to the parent. Sum of euclidean distance between the higher density objects and lower density objects. Lower density object should cover at least 80\% of higher density object. Algorithm \ref{alg:mergeScore} describes the process of merge score. This process is repeated for all the segmented regions at every selected hierarchical level. Hierarchical clustering gives a parent-child relationship of clusters also, we can use this relationship to avoid unacceptable merging of objects. Objects having at least 3 prestige score from each level are up-sampled to full resolution image. Result of merging process is shown in Figure \ref{fig:detectedObjects}, where \ref{fig:foundClusters} represents the detected ROIs and \ref{fig:mergeClusters} shows the merged objects.

\begin{algorithm}
\caption{Merge Score}
\label{alg:mergeScore}
\KwResult{Merge Score}
 \textbf{Given}\;
 Labels = \{$L_1, L_2$,.....,$L_n$\}\;
 
 \For {i = 1 to n}{
 	currentLabel = \{$L_i$\}\;
 	Objects = Object by current current label\;
 	numObjects = number of Objects by current label\;
 	\For {j = 1 to numObjects} {
 	mergeScore[i][j] = 1;\;
 	}
	dist = distanceL2 (Objects[i], Objects[i-1] )\;
	\If{dist \textless 0.2}{
		mergeScore[i][j] += mergeScore [i-1][j] \; 
	}	
}

\end{algorithm}

\subsection{Features for False Positive (FP) Analysis} 
Following set of features are extracted to classify objects into true mass and breast tissue (false positive). These features are well-established statistical features and finalized by radiologist too after analyzing the prominent patterns of masses on mammograms.

\subsubsection* {Region Contrast:}	 Generally, mass is imprinted on mammogram as a dense object as compared to its surroundings, having at least a uniform density. We used this property for classification between true mass and breast tissue. Region Contrast is computed as a difference between mean intensities of foreground and background in ROI. Foreground area is the selected mass or object while background represents the background area surrounding this object. Regions which results in negative values of region contrast are rejected for further processing.
\subsubsection* { Mean Gradient:} Gradient monitors the directional change in intensity. Gradient magnitude describes that how quick the image is changing. We calculated the mean gradient of the boundary pixels which strengthens the compactness of the region( described later). 
\subsubsection* {Entropy:} The concept of entropy is in information theory which states the probabilistic behavior of the information sources. This statistical measure is a measure of randomness that is used to characterize the texture of image.
\subsubsection* {Standard Deviation:} It is popular term in statistics which gives a measure of spread of data. This represents the measure, that how much close the points are in the given region of the image.
\subsubsection* {Compactness:}  The value of compactness gives the ratio of contour which encloses an area. it is defined as:
\begin{equation}
compactness\ =\ 1-{{4*pi*A}\over{P^2}}
\end{equation}
where  A is Area of object enclosed by perimeter P. Usually Benign masses have higher value of compactness, because it defines that small perimeter is enclosing a bigger area. We have used this feature in benign vs malignant classification too.

\subsection{Classification Model}
SVM(Support Vector Machine) was used to classify the masses. We selected support vector machine as it gives good results for binary classification. The basic idea behind SVM is to separate the input data by optimal method. As our data is not linearly separable, we used Gaussian RBF (Radial basis function) kernel. Sigma and C are two important factors for RBF kernel. optimal values for RBF were grid-searched between $10^{-3}$ to $10^3$. Harmonic Mean (HM) is calculated to compare the C and sigma pairs. Harmonic Mean is defined as:
\begin{equation}
\label{eq:hmean}
HM\ =\ {{2*\ sens*spec}\over{sens+spec}}
\end{equation}
where sens is sensitivity and spec represents specificity of the system.
We adopted a 10-fold cross validation technique to train,test and validate the data.

\section{Results and Discussion}
\label{results and discussion}
\subsection{Image Database}
This study was carried out on images from two databases. We selected 56 images from mini-MIAS database (\cite{suckling1994mammographic}). It includes 13 normal, 13 malignant and 30 benign cases. The dataset include all types of masses from both classes (benign and malignant). Table 1 shows the overview of number of cases used in experiments from MIAS-dataset.
We also selected 76 cases from DDSM database (\cite{heath1998current}:\cite{heath2000digital}). Table 2 shows the summary of DDSM database \par
\begin{table*}
\caption{MIAS (Mammographic Image Analysis Society) dataset}
\label{tab:miasdataset}
\centering
\begin{tabular}{|c|c|c|c|c|c|}
\hline
\multicolumn{3}{|c|}{\textbf{Benign}} &
\multicolumn{3}{|c|}{\textbf{}Malignant} \\
\hline 
\textbf{Dense} & \textbf{Fatty-Glandular} & \textbf{Glandular} & \textbf{Dense} & \textbf{Fatty-Glandular} & \textbf{Glandular} \\ \hline
8 & 12 & 10 & 2 & 6 & 5 \\
\hline
\end{tabular}
\end{table*}
\begin{table}
\small
\centering
\caption{DDSM (Digital Database for Screening Mammography) dataset information}
\label{tab:ddsmdataset}
\resizebox{8.5cm}{3.3cm}{
\begin{tabular}{|l|l|c|}
\hline
\textbf{Property} & \textbf{Description} & \textbf{Image Count} \\ 
\hline
\multirow{4}{*} {\textbf{Density}} & 1 & 14 \\ \hhline{~--}
&2&43\\ \hhline{~--}
&3&34\\ \hhline{~--}

&4&6\\
\hline

\multirow{7}{*} {\textbf{Shape}} & Fine Linear Branch & 2 \\ \hhline{~--}
&Irregular&16\\ \hhline{~--}
&Irregular Architecture&8\\ \hhline{~--}
&Lobulated&9\\ \hhline{~--}
&Oval&8\\ \hhline{~--}
&Pleomorphic&4\\ \hhline{~--}
&Round&3\\ 
\hline

\multirow{9}{*} {\textbf{Margin}} & Circumscribed & 10 \\ \hhline{~--}
&Circumscribed ill Defined&1\\ \hhline{~--}
&ill Defined&9\\ \hhline{~--}
&ill Defined Spiculated&1\\ \hhline{~--}
&Microlobulated&1\\ \hhline{~--}
&Obscured&3\\ \hhline{~--}
&Obscured ill defined&2\\ \hhline{~--}
&Obscured ill defined spiculated&3\\ \hhline{~--}
&Spiculated&15\\ 
\hline

\multirow{3}{*} {\textbf{Pathology}} & Benign & 14 \\ \hhline{~--}
&Malignant&38\\ \hhline{~--}
&Normal&46\\ 
\hline
\end{tabular}
}
\end{table}
\subsection{Detection of ROIs}
Our proposed preprocessing steps detected almost all masses in the dataset. Through careful examination of ROIs, we found that our algorithm missed two cases in MIAS database. One from Malignant and the other from Benign case (mdb179 and mdb191), Dense-glandular and Fatty Glandular. The contrast in these two images was very high and distributed, making it difficult to detect isolated regions. All other masses were successfully detected. This results in the detection accuracy of 95.3\%. The detection accuracy on DDSM dataset was 97.3\%. We missed 2 cases. Detected ROIs were carefully compared with the given ground truth data.
\begin{figure*}
\begin{subfigure}{.5\textwidth}
\includegraphics[,height=5cm,keepaspectratio]{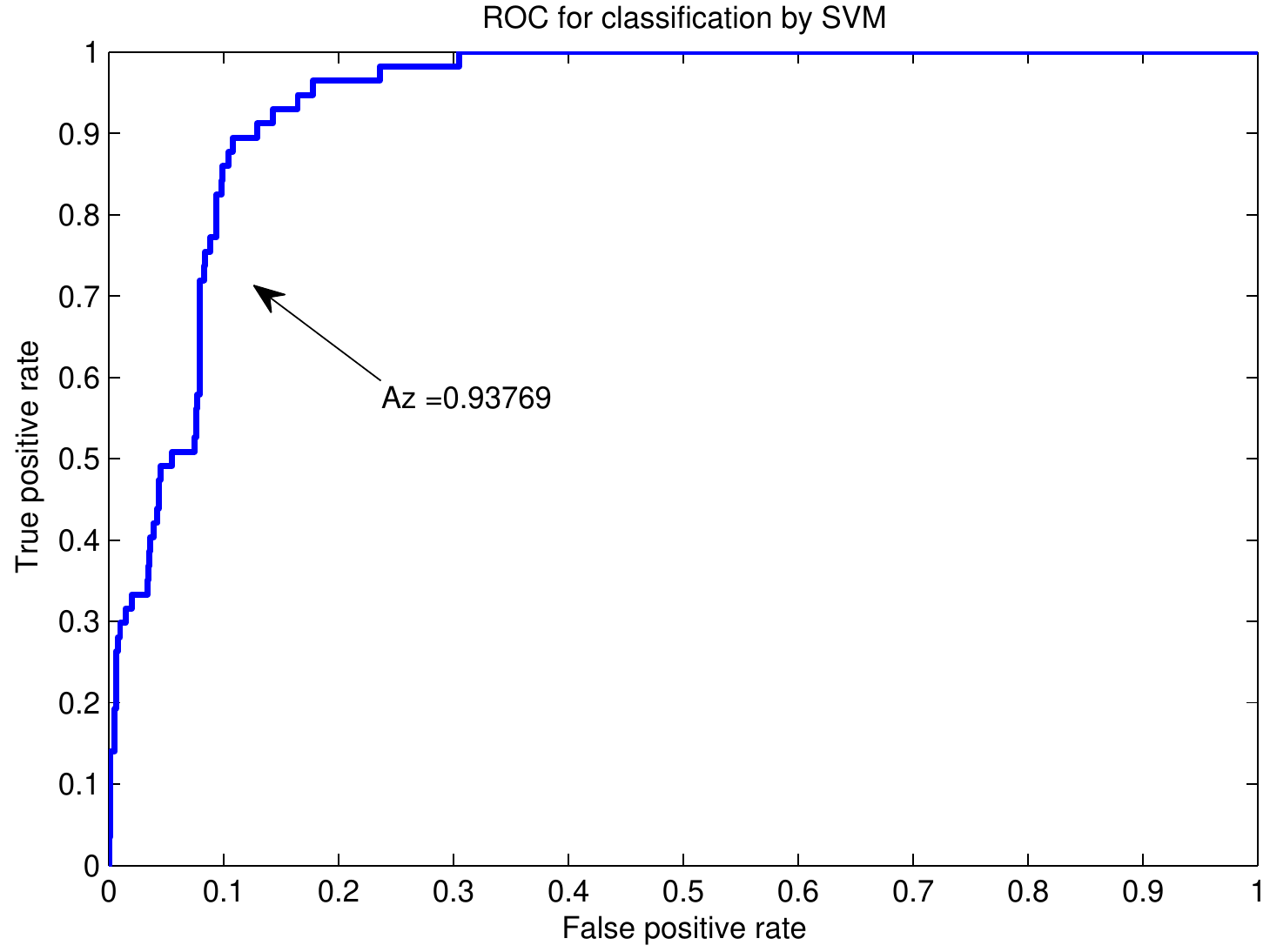}
\caption{ROC of Normal and Mass (MIAS dataset)\label{fig:miasNormal}}
\end{subfigure}
\begin{subfigure}{.5\textwidth}
\includegraphics[,height=5cm,keepaspectratio]{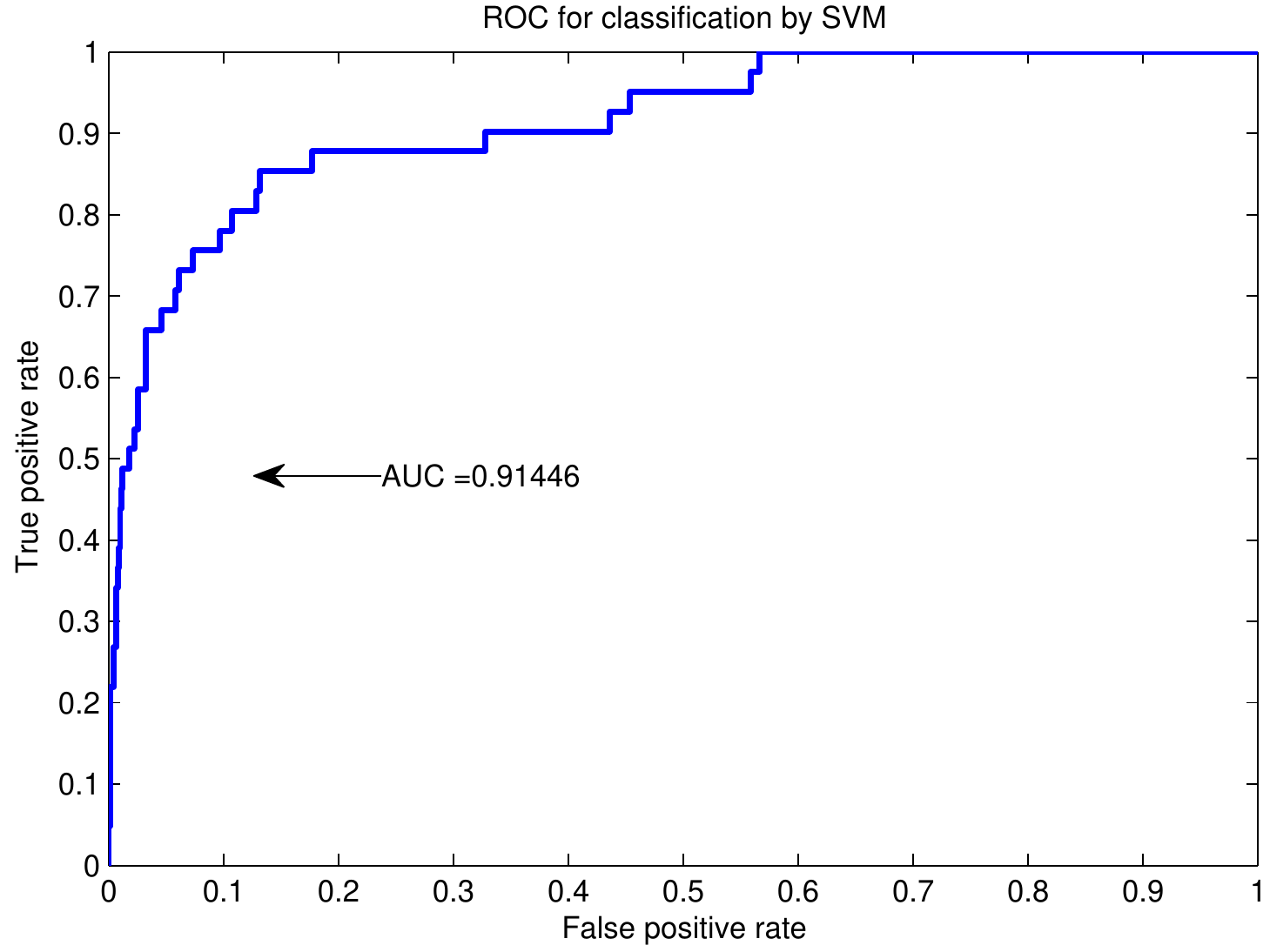}
\caption{ROC of Normal and Mass (DDSM Dataset)\label{fig:ddsmNormal}}
\end{subfigure}
\begin{subfigure}{.5\textwidth}
\includegraphics[,height=5cm,keepaspectratio]{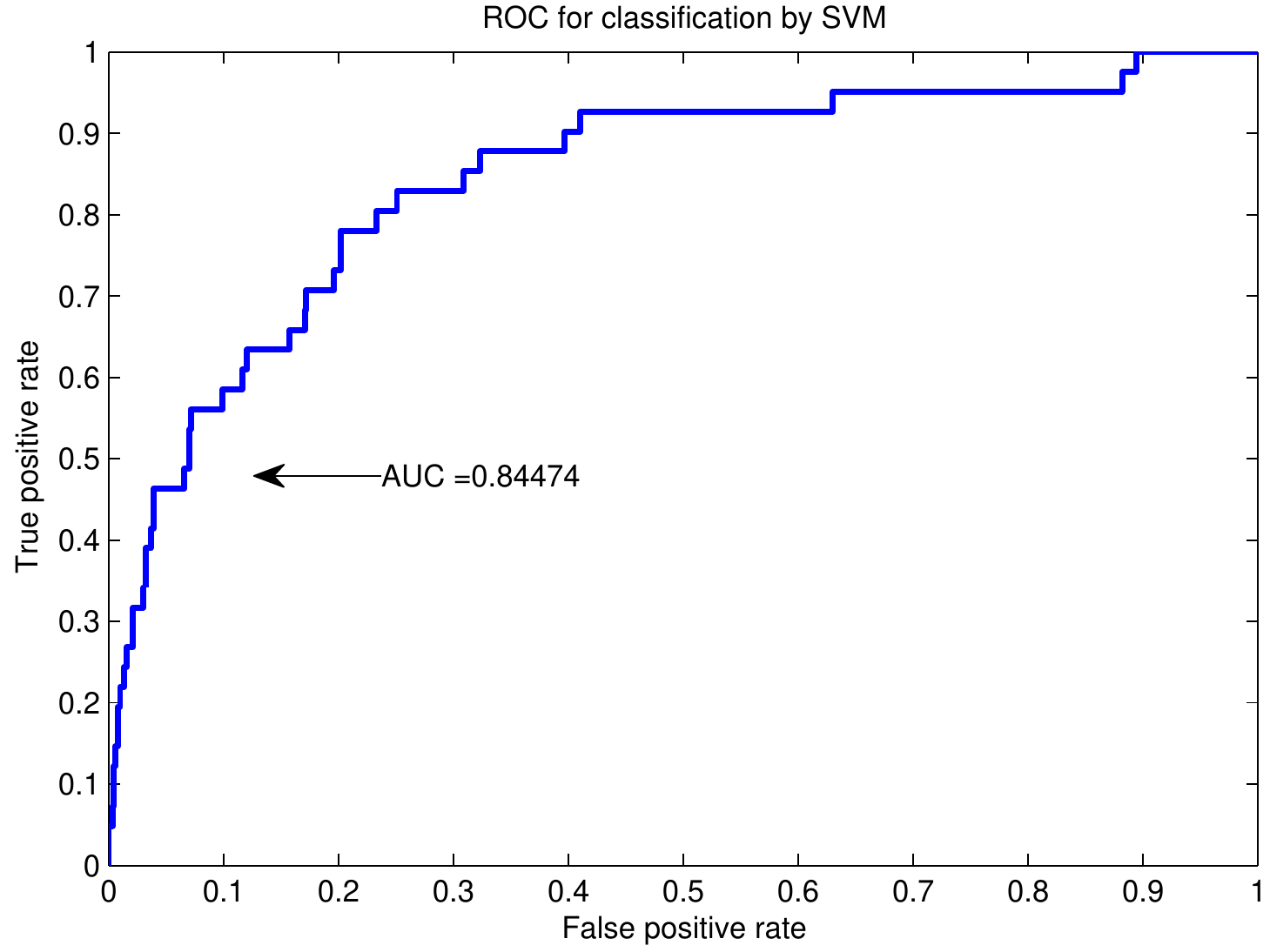}
\caption{ROC of training on MIAS and Test on DDSM \label{fig:trainMiasTestDDSM}}
\end{subfigure}
\begin{subfigure}{.5\textwidth}
\includegraphics[,height=5cm,keepaspectratio]{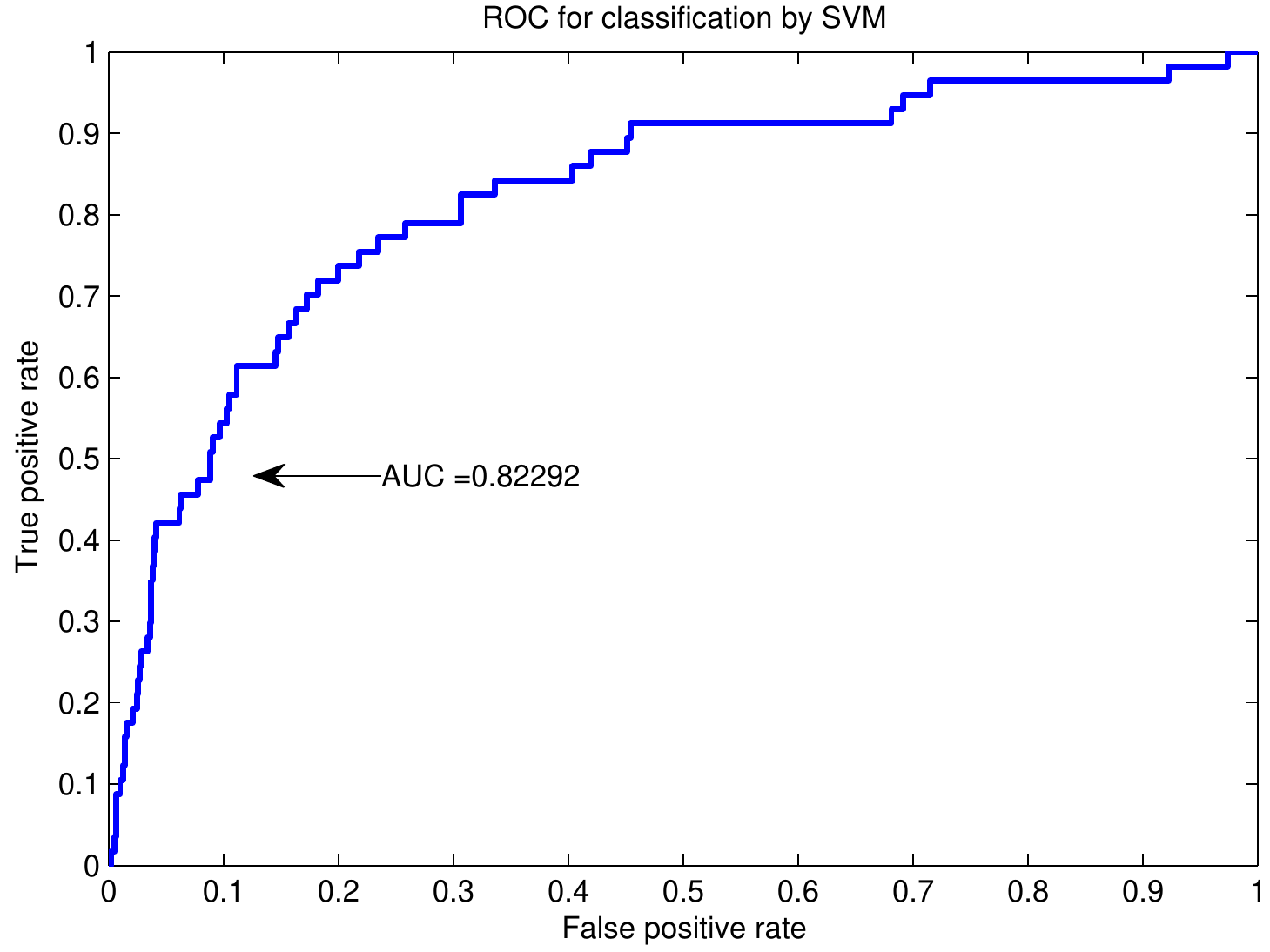}
\caption{ROC of training on DDSM and Test on MIAS \label{fig:trainddsmTestmias}}
\end{subfigure}
\begin{subfigure}{.5\textwidth}
\includegraphics[,height=5cm,keepaspectratio]{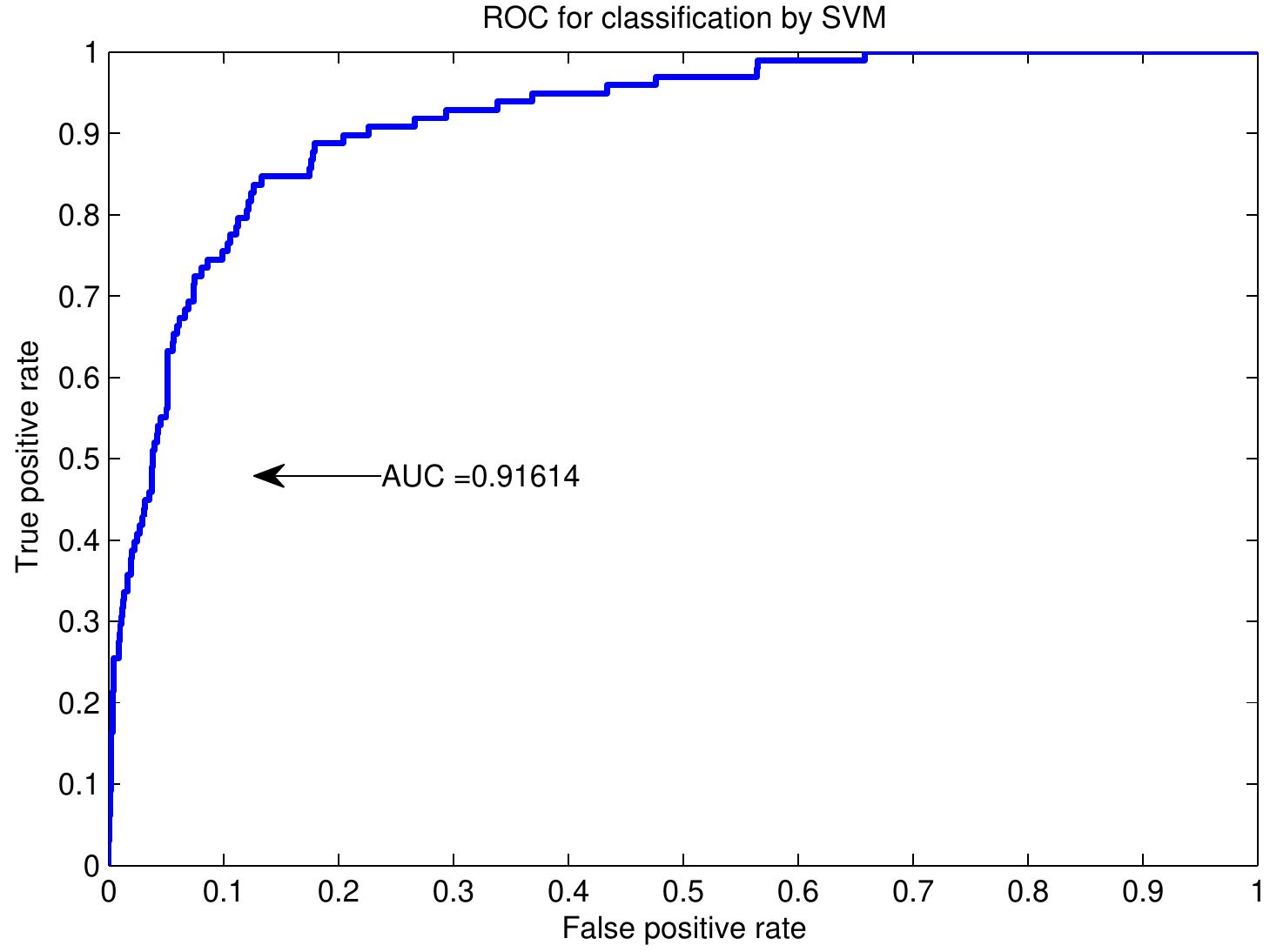}
\caption{ROC of MIAS and DDSM combined \label{fig:miasplusddsm}}
\end{subfigure}
\caption{Breast tissue vs Mass classification results (ROC plots) }
\label{fig:NormalvsMass}
\end{figure*}
\subsection{Normal and Mass Differentiation}

\begin{figure}
\centering
\includegraphics[scale=0.1]{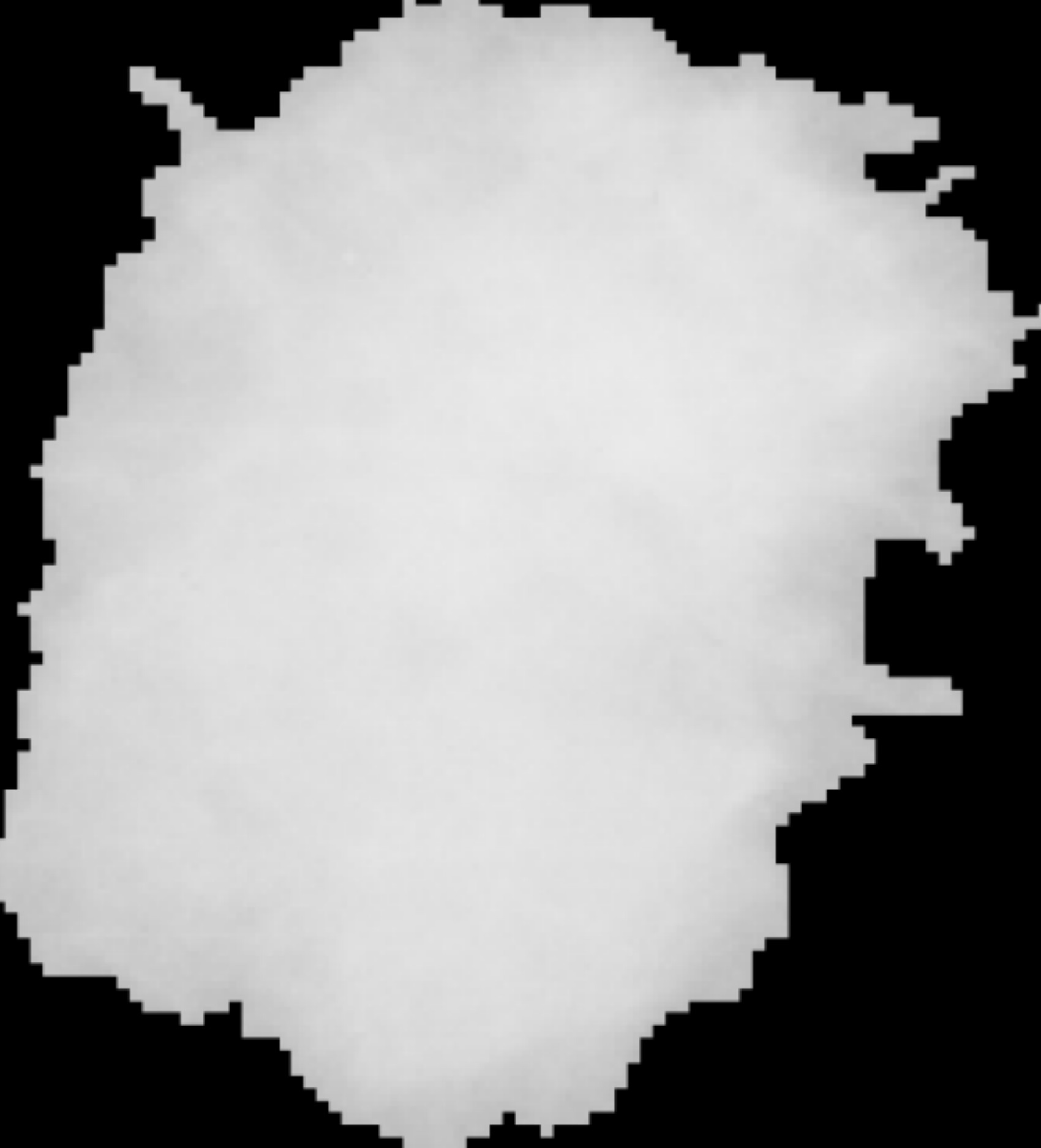}
\caption{Detected Mass ROI \label{fig:roi}}
\end{figure}

Our algorithm detected all the malignant masses except one (mdb0186) on MIAS dataset. However we did not get so prominent success on benign masses. 30 cases were tested but Algorithm failed to detect 6 masses. Three of these missed masses were Fatty (mdb069, mdb080 and mdb195), two were Dense-glandular (mdb193 and mdb290) and one was Fatty-glandular (mdb190). The total accuracy of the system was 83.43\%. Figure \ref{fig:roi} shows the example ROI which is classified as mass.

We further investigated the missed cases and found the following observations. In the first missed case (mdb069), the margin and boundary with wide transition zone, if we compare with opposite side breast, the lesion could be detectable, and in  clinical practice, we describe it as architectural distortion. In case of mdb080, the tumor lesion is subtle ill margined, non-mass like parenchymal asymmetric pattern. In case of mdb195, the malignant lesion is almost isodensed to the normal breast fatty parenchyma. So the detection is not feasible. In mdb186 we found that the mass has poor contrast and also it lacks the dense region. It\'s contrast with respect to the surrounding was very poor. Benign cases, where algorithm was unable to classify masses, we observed that, in three fatty and one fatty glandular case (mdb069, mdb080, mdb190 and mdb195) the masses were not clear. They do not have center core region and their contrast with respect to their surrounding was poor too. We are confident that if we add some good contrast enhancement technique, our algorithm performance will be improved by classifying above described cases as well. The remaining two dense-glandular cases (mdb193 and mdb290) do not follow the assumption we made in this paper (they do not have glowing effect), so features values were not good in these cases to classify them. To successfully detect masses in these cases, it may require additional methods or include more features. In the present work, we did not reject any region because of its size, this results in generating a large number of false positives. Although our classification phase reduces the number of FPs, but we aim to reduce the number FPs by improved algorithm in future work. We also believe that automatic breast density assessment before applying our method will improve the performance (\cite{law2014automated}). 
\par We validated the results by plotting the receiver operating characteristic (ROC) curve, which illustrates the performance of binary classifier system as its discrimination threshold is varied. Figure \ref{fig:NormalvsMass} shows the ROC curve of classification between normal and mass data, which is obtained by varying the threshold on the probabilities by classifier (SVM). AUC refers to the Area Under Curve. Table \ref{tab:massvsnormal} shows the classification results in terms of specificity and sensitivity.In medical domain, only sensitivity is not important, algorithm should yield good specificity results also. As previously described, we used harmonic mean (equation \ref{eq:hmean}) to get the best pair of specificity and sensitivity
\par Algorithm missed 2 cases from malignant category and 6 from benign category of DDSM dataset. The maximum sensitivity and specificity pair we achieved is 91.32\% and 85.05\% respectively. Average sensitivity and specificity is 76.19\% and 87.05\% respectively.

\par We also tested our algorithm for its generality by training it on one dataset and testing on the other. Algorithm was trained on MIAS dataset, tested on DDSM and vice versa. Algorithm results in table \ref{tab:massvsnormal} confirm our claim that proposed algorithm is not limited to some limited type of masses or abnormalities. It covers a wide spectrum of masses. Distribution of the dataset is uneven, which degrades the performance of learning algorithm. 

\begin{table}[]
\centering
\caption{Specificity and Sensitivity of Mass vs Normal classification}
\label{tab:massvsnormal}
\begin{tabular}{|l|l|l|l|}
\hline
\textbf{Training} & \textbf{Testing} & \textbf{\begin{tabular}[c]{@{}l@{}}Senstivity\\ (\%)\end{tabular}} & \textbf{\begin{tabular}[c]{@{}l@{}}Specificity\\ (\%)\end{tabular}} \\ \hline
MIAS              & MIAS             & 90.47                                                              & 82.95                                                               \\
DDSM              & DDSM             & 78.03                                                              & 87.02                                                               \\
MIAS+DDSM         & MIAS+DDSM        & 75.51                                                              & 84.12                                                               \\
MIAS              & DDSM             & 87.80                                                              & 64.88                                                               \\
DDSM              & MIAS             & 75.51                                                              & 84.12 \\
\hline                                                              
\end{tabular}
\end{table}

\par Investigation of the missed cases confirms the reasons described earlier. Case0004 from DDSM shows poor contrast around the mass, making it difficult to be detected. Case0005, case0006, and case008 does not follow the assumption we made in the paper. More features may be required to detect those masses. We also calculated the number of false positives per image which was 4.67 FP/Image. This number is calculated only on Normal Images to give the fair view of the system.
\subsection {Comparison with existing algorithms}
\cite{rodriguez2013detection} stated their results of mass detection phase they achieved 84.4\% detection accuracy. Their algorithm is based on image enhancement and after that Gaussian Markov Random Field (MRF) is used for mass segmentation. They did not classified the ROIs into mass and non-mass regions. \cite{kuo2014mass} also reported their detection accuracy as 94.44\%. They presented a particle swarm optimization (PSO) based detection technique. Our algorithm outperformed previously reported detection accuracies. 

\par Work presented by \cite{garma2013detection}, \cite{torrents2014breast}, \cite{eltoukhy2014optimized}, and \cite{choi2012multiresolution} can be consider as the baseline in recent work on this domain. \cite{choi2012multiresolution} implemented a fully automated system. They extracted local binary pattern LBP features and the classification is done by SVM. They also proposed a feature selection technique.  \cite{choi2012multiresolution} reported their performance in terms of sensitivity and 75.86\% is reported for overall CAD performance on MIAS database.  \cite{gardezi2014analysis} reported their results on already selected 305 ROIs and achieved a sensitivity of 76.53\%. They extracted features from Grey-level co-occurrence matrices (GLCM) and then classify features into mass and non-mass regions. \cite{eltoukhy2014optimized} proposed the technique of curvelet transformation, feature selection and then classification by SVM. They manually cropped the ROIs and then applied their algorithm. Their reported accuracy is higher 90\%, but their algorithm is not fully automated, they lack mass detection phase.
All methods were tested on separate dataset, cross validation between the datasets was never performed. 

\section{Conclusion}
\label{Conclusion}
This paper proposes a new mass detection in mammogram images. The proposed method is fully automated. It finds the candidate regions by segmenting the salient regions in mammogram and then extract features to differentiate between breast tissue and mass.Promising results are obtained in mass identification and normal vs mass tissue classification. Classification results confirms that the segmentation process extracts enough information to find masses and localize it in mammogram. Experiments were performed on mini-MIAS and DDSM databases to show the usefulness and generalization of the proposed algorithm. Correlating the full image set (CC and MLO) is considered as future work that can help to identify the architectural distorted mammograms also.

\bibliographystyle{abbrv}
\bibliography{refs}  

\end{document}